\documentclass[letterpaper, 10 pt, conference,twocolumn	]{ieeeconf}  

\IEEEoverridecommandlockouts                     
\usepackage{graphics} 
\usepackage{epsfig} 
\usepackage{mathptmx} 
\usepackage{amsmath} 
\usepackage{amssymb}  
\usepackage{algorithm}
\usepackage{algpseudocode}
\usepackage{xcolor}

\usepackage[backend=biber,sorting=none]{biblatex} 
\bibliography{ref}

\title{Floating-base manipulation on zero-perturbation manifolds}

\author{Brian A. Bittner, Jason Reid, Kevin C. Wolfe
\thanks{Johns Hopkins University Applied Physics Laboratory, Laurel, MD, USA.
        {\tt\small brian.bittner@jhuapl.edu}}
}
\begin{document}

\maketitle
\thispagestyle{empty}
\pagestyle{empty}

\begin{abstract}
To achieve high-dexterity motion planning on floating-base systems, the base dynamics induced by arm motions must be treated carefully.
In general, it is a significant challenge to establish a fixed-base frame during tasking due to forces and torques on the base that arise directly from arm motions (e.g. arm drag in low Reynolds environments and arm momentum in high Reynolds environments).
While thrusters can in theory be used to regulate the vehicle pose, it is often insufficient to establish a stable pose for precise tasking, whether the cause be due to underactuation, modeling inaccuracy, suboptimal control parameters, or insufficient power.
We propose a solution that asks the thrusters to do less high bandwidth perturbation correction by planning arm motions that induce zero perturbation on the base.
We are able to cast our motion planner as a nonholonomic rapidly-exploring random tree (RRT) by representing the floating-base dynamics as pfaffian constraints on joint velocity. These constraints guide the manipulators to move on zero-perturbation manifolds (which inhabit a subspace of the tangent space of the internal configuration space).
To invoke this representation (termed a \textit{perturbation map}) we assume the body velocity (perturbation) of the base to be a joint-defined linear mapping of joint velocity and describe situations where this assumption is realistic (including underwater, aerial, and orbital environments).
The core insight of this work is that when perturbation of the floating-base has affine structure with respect to joint velocity, it provides the system a class of kinematic reduction that permits the use of sample-based motion planners (specifically a nonholonomic RRT). 
We show that this allows rapid, exploration-geared motion planning for high degree of freedom systems in obstacle rich environments, even on floating-base systems with nontrivial dynamics.
\end{abstract}

\section{Introduction}

\begin{figure*}
\label{fig:eetracking}
\includegraphics[width=\textwidth]{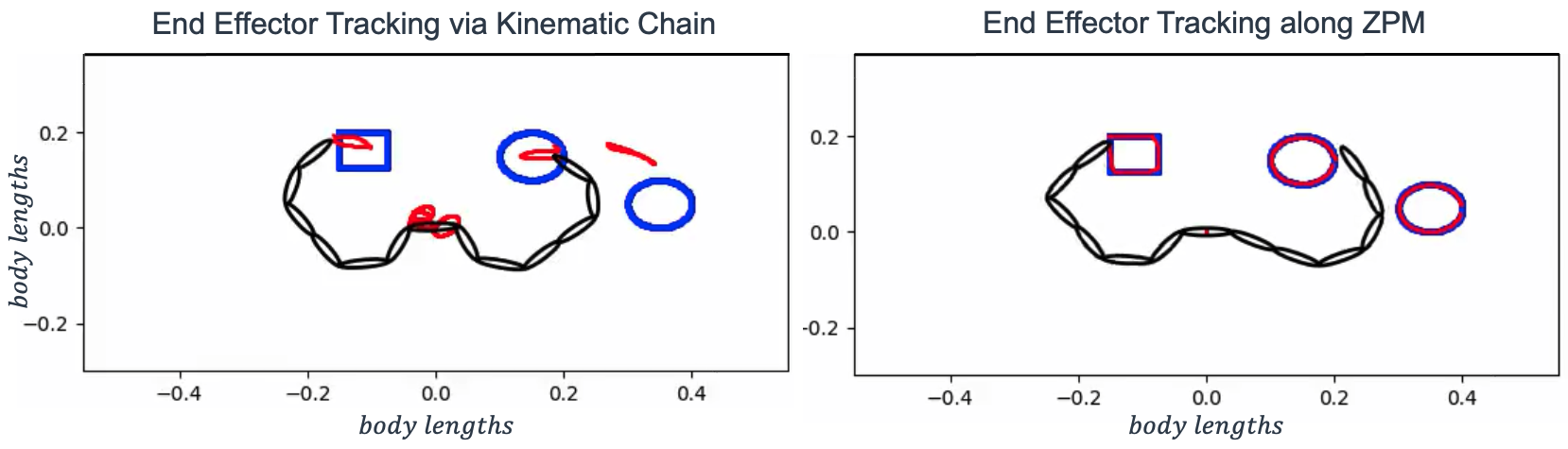}
\caption{A thirteen-link Purcell swimmer with knowledge of its kinematic chain via (\ref{eq:kin}) (left) is unable to track (red proximal to end effectors) the shapes (blue) sequentially attempted right to left due to perturbation induced on the base (red proximal to base) during tracking. The thirteen-link Purcell swimmer with knowledge of its dynamics as well as its kinematic chain written as pfaffian constraints via (\ref{eq:dynpfaffian}) (right) is able to succeed given the same task. 
This tracking result is explained in detail in Section \ref{sec:tracking}.
The performant tracker leverages a surface in the joint velocity space which we term a zero-perturbation manifold (ZPM). This surface is the core element of a new planning framework covered in Section \ref{sec:method}. Accompanying video in Supplementary Materials.}
\end{figure*}

Floating-base manipulation platforms offer the potential to perform critical tasking in underwater \cite{kinsey2006survey,simetti2014floating,cieslak2015autonomous,brantner2021controlling}, aerial \cite{ruggiero2018aerial,kobilarov2014nonlinear,zhao2022versatile}, and orbital \cite{kazanzides2021teleoperation,papadopoulos2021robotic} environments. 
While manipulation has succeeded in high complexity environments \cite{krotkov2018darpa,mahler2019learning,mcconachie2020manipulating}, floating-base systems have yet to achieve the same ability to precisely manipulate objects in similar environments.
The inability to physically anchor any link of the kinematic chain can induce complex dynamics on the ungrounded robot.
These dynamics present differential constraints that are often challenging for the motion planner to accommodate when pursuing precision tasking.
If the floating-base manipulator assumes a kinematic operating environment, perturbations on the base can cause imprecision in movement that is unpredictable and unsafe, especially in obstacle rich environments.
By considering these perturbation dynamics, one can pursue more precise positioning of the end effector(s). 

Typically in floating-base manipulation, the dynamics are represented as a second-order mechanical system. 
Such a representation stipulates that given a path in the joint configuration, the rate along that path has a variational effect on the evolution of the robot in its workspace.
These timing-based concerns induce substantial complexity on the motion planning problem; these planning problems are classified as kinodynamic.
Generally, these kinodynamic path planning problems are formulated as trajectory optimization problems. 
These approaches typically approximate the Hamilton-Jacobi-Bellman equation, which recursively formulates the optimal control law for a physical system.
These exploitative methods excel at finding local refinements to seeded trajectories that leverage the agile capabilities of the system \cite{kleff2021high,rybus2017control,lee2020aerial}, as long as the underlying model maintains high fidelity to the realized system and environment.
A separate trajectory optimization approach involves policy gradient methods \cite{ratliff2009chomp,kalakrishnan2011stomp}, 
which are performant at rapidly finding collision free paths, but again are typically used to find local refinements to approaches (due to the exploitative nature of gradient descent).\footnote{While stochastic formulations are designed to specifically avoid local minima, we emphasize that the nature of a policy gradient search is far more localized than sampling-based path planning methods.}
Additionally, this class of solutions suffer from poor sampling scalability with the dimension of the configuration -- termed the \textit{curse of dimensionality}.
This notion of scalability is encountered in many practical scenarios, such as adding joints to improve the reachable configurations of a robot in key task domains (e.g. pick and place in a cluttered environment). 
Exploitative trajectory optimization methods thus tend to be insufficient for accomplishing many critical tasks due to their locality and sample inefficiency. 
However, as is typically the case in mobility architectures, a higher level planner might successfully hand off a performant, coarse plan intended for final refinement by the trajectory optimizer.

The main contribution of this work is the justification, design, and implementation of a framework that decouples perturbations during floating-base manipulation within a sampling-based motion planner (specifically an RRT).
We show that a kinematic reduction of the floating-base dynamics can be applied to systems through mechanical design, environment selection, and thruster-aided pose control. 
This reduces the motion planning problem from kinodynamic to kinematic, making sampling-based motion planning algorithms accessible. 
Empirically, such methods extend to high dimensional problems with practical sample efficiency \cite{elbanhawi2014sampling}. 
Sample efficiency for large dimensions is critical, since as described earlier, higher numbers of joints will be required for complex tasking in cluttered environments. 
Gaining this property (sample efficiency for platforms with significant internal complexity -- required for many critical application domains) while respecting the dynamics is the core capability that we pursue through this work. 
Using the method detailed in Section~\ref{sec:method}, we structurally expect (A) higher precision end effector tracking and placement than planners with no dynamics knowledge as well as (B) rapid computability of valid plans in many-joint, obstacle rich scenarios. 
We show in obstacle rich scenarios that for emplacement and extraction tasks, the proposed method can shift induced perturbation to a trivial moment during the plan, allowing the ZPM to fascilitate a high precision engagement with a target configuration.

The class of kinematic systems we assume stipulates that the body velocity is a joint-defined linear mapping of joint velocity. 
These pfaffian constraints can be used at any configuration to identify a subspace of joint velocities where instantaneous joint motion yields zero perturbation (a sufficient condition being that the dimension of the internal configuration is greater or equal to the dimension of the position space). 
These pfaffian constraints are termed a \textit{perturbation map}, and this map provides a nonheuristic manifold that can be used to guide the system to rapidly find collision-free plans. 
No objective function designing or tuning is intended to be a part of implementing the methods motivated in this work.

In Section~\ref{sec:bg}, we provide background on kinematic representations of floating-base systems in application-oriented physical scenarios. 
We then define and discuss key properties of the ZPM in Section~\ref{sec:zpm}, followed by an example of an end-effector tracking shapes without inducing motion on the base in Section~\ref{sec:tracking}.
In Section~\ref{sec:method} we cover the proposed method for kinematic planning on floating-base systems followed by Section~\ref{sec:results} where we demonstrate global path planning 
using the proposed method.
Notably, the method outperforms (A) kinematic algorithms that do not leverage dynamical information and 
(B) trajectory optimizations that find locally optimal, low performance solutions.
Finally, Sections~\ref{sec:discussion} and \ref{sec:conclusion} include discussion and conclusion of the results.

\section{Background: kinematic reductions of Lagrangian systems}
\label{sec:bg}

In this work, we seek to compute plans for floating-base systems when their motion model can be written as a first-order (kinematic) system expressed as pfaffian constraints connecting velocity on the shape space (joint space) $B$ to velocity on the position space $G$.
These constraints are expressed as
\begin{equation}
\label{eq:pertmap}
\dot{x}_{b} = P(\theta)\dot{\theta}
\end{equation}
yielding a linear mapping, $P$ (as a function of $n$ joints $\theta \in B$) from joint velocity $\dot{\theta} \in TB$ to body velocity $\dot{x}_b \in TG$.
This structure will be shown to be an exact representation of the physics in conventional orbital environments and scenarios governed by Rayleigh dissipation. 
We'll show that violations of this physical structure can be related to intuitive phenomena and suggest that in many cases these complexity inducing features can be designed out of the problem through mechanical design, environment selection, and control.

\subsection{Obtaining Environmental Symmetry}
Gravity directly violates environmental symmetry, inducing position and orientation dependent forces on the floating-base.
This will, for instance, cause the robot to move when the joints are still, clearly violating our assumed model at $\dot{\theta}=0$.
Gravity is however negligible for orbital robots as well as neutrally buoyant underwater systems\footnote{Some platforms such as RE2s sapien class system are designed to be neutrally buoyant to avoid dynamical complexity.}. In less convenient scenarios, one might supply a model-based control law to compensate for gravity with thrusters, through which one could regain environmental symmetry of the floating-base under this thruster-governed control law.

When we can assert symmetry of the dynamics over $G$ for Lagrangian systems via a powerful reframing of the dynamics that provides three distinct equations written for the generalized momentum, body velocity, and joint acceleration of the system.
This result from the field of geometric mechanics is known as the reduced Lagrangian \cite{ostrowski1998geometric}.

\subsection{Mitigating Interaction Between Inertia and Friction}

When the generalized momentum starts at zero (as is the imagined case for many floating-base tasks), the dynamics of the body are governed exclusively by conservation of zero net angular and linear momentum.
The resulting dynamics take the form of (\ref{eq:pertmap}) and is referred to as the \textit{mechanical connection} \cite{marsden1998symmetries}.
A similar phenomena occurs when the motion model becomes dominated by Rayleigh dissipation.
Here, friction-governed forces applied along the body yield a viscous interaction with the fluid, again identical to the form in (\ref{eq:pertmap}) and termed the \textit{viscous connection} \cite{kelly1995geometric}.

Coupling of forces driven by inertia and friction result in second-order behaviors. 
Take the free-floating astronaut as a \textit{mechanical connection}. 
A brief friction-driven interaction of brushing a satellite imparts an uncancellable impulse on the astronaut. 
Post interaction it has obtained a net momentum that it cannot cancel through internal motion, providing a drift that violates the first order structure of the mechanical connection.
Take a micro-swimmer governed by a \textit{viscous connection}. 
If we iteratively increase the inertia of the swimmer, it will eventually noticeably drift as a function of accumulated momentum while swimming. 
This drift again violates the first order structure of the \textit{viscous connection}.

Systems can be engineered to reduce interaction of these forces. 
Underwater robots might move slowly to reduce excitation of momentum and leverage the structure of a drag-dominated environment (assuming gravity compensation).
Aerial robots might be designed to avoid arm-related drag to encourage isolation of momentum driven dynamics (assuming gravity compensation).
Orbital robots experience no environmental friction unless contacting objects in space.

\subsection{Approximating Nonviscous Friction}

In general, friction can certainly act in non-viscous ways, violating the structure of (\ref{eq:pertmap}). 
One can think of the viability of the Rayleigh dissipation model as the viability of a first-order Taylor series approximation of the more general friction-governed motion model $\dot{x}_b | \theta = f(\dot{\theta})$. 
\footnote{In many cases this can be a complex, non-smooth function (e.g. in the case of coulomb friction) motivating the use of stochastic linearization to obtain useful model parameters. Isotropy of the friction can also have a significant impact on behavior \cite{ma2018friction,alben2019efficient}, further motivating the use of stochastic linearizations to build linearizations that support prediction over a desired range of velocities.}
However, it has been observed in practice that the assumption of a Rayleigh dissipation model provides a strong fit across a surprisingly broad class of robots \cite{astley2020surprising,zhao2022walking}.

In this section we have detailed the motion model of interest and the relevant assumptions for dynamics and control to arrive at its structure. 
We now outline a method which allows us to use sampling-based motion planning while respecting a broad class of nontrivial floating-base dynamics.

\section{Zero Perturbation Manifolds}
\label{sec:zpm}

We now shift focus to explaining how the representations motivated in Section~\ref{sec:bg} populate a manifold that can be exploited for motion planning that induces zero perturbation on the floating-base.
The linear mapping $P(\theta)$ has a null space computable at all shapes $\theta$ 
\begin{equation}
null(P(\theta)) \subset TB 
\end{equation}
which inhabits a subspace of the shape velocity space $TB$.
This null space provides a basis in the joint velocity space for exploration from a shape that will infinitesimally invoke zero perturbation on the base.
This null space changes as a function of $\theta$, and so during a plan, the allowable motions available to the joints are subject to change.
When the null space changes continuously with respect to shape, the null space populates a connected submanifold of joint velocity space. 
We know this since the notion of connectedness is preserved under continuous mappings.
However, the null space is provably not continuous in this sense at singular $P$ where the rank of the nullspace increases over that of non-singular $P$.
Thus, the zero perturbation manifold (ZPM), $null(P)$, has a basis of size $dim(B)-dim(G)$ and continuously changes over its connected regions.
It is however a larger manifold that is not continuous, e.g. at locations of singular $P$.

Continuous or not, the null space indicates at all internal configurations instantaneous directions to move the joints that induce zero base perturbation.
Continuity of the null space is however important for generating $C^1$ smooth arm motions. 
If a null space basis vector disappears while moving a joint in that direction, it will appear as a nonsmooth path in joint space (if persisting along the zero perturbation manifold).
This manifold is not unlike leveraging the null space of components of the manipulator Jacobian to stabilize elevation, roll and pitch when moving a weight along a table, as is done in work that in-part inspired this approach \cite{berenson2009manipulation}.

In many classical manipulation scenarios, additional joints provide exploitable redundancy. Likewise, additional joints directly increase the dimension of the zero perturbation manifold. We make a final note about the challenge of connecting two specific points in the shape space. 
The probability of the null space directly aligning with two randomly selected shapes is approximately zero:
\begin{align}
\label{eq:hard_to_connect}
\forall \delta \in B: \textbf{p}(null(P(\theta)))^T null(P(\theta+\delta)=1)\approx0.
\end{align}
where $\textbf{p}$ provides our single-case use of notation for probability.
While it is easy to grow a tree along the null space from a known joint location, it can be hard to grow the tree along the null space toward a known joint location.
One might think of an Ackermann steering vehicle (a system whose maneuvers are governed by pfaffian constraints) that is placed arbitrarily close to a final location, then asked to arrive at that location directly, without knowledge of Lie brackets or "parallel parking" maneuvers. 
In many scenarios, there is not a direct route to that location.
In higher-degree of freedom floating base manipulation problems, it is almost never the case that such as direct route exists.
To our knowledge, it is an open problem whether structures exist to procedurally connect such proximal shape configurations along the zero perturbation manifold. 
Lie brackets offer an interesting avenue for further inspection, but are not the subject of this work.

\section{Local End Effector Tracking along the Zero Perturbation Manifold}
\label{sec:tracking}
Here we leverage the exploitable redundancy property discussed in Section~\ref{sec:zpm} to show that a many jointed, bimanual swimming robot can complete a welding task with a closed form solution. 
This solution will implicitly leverage the structure of the zero perturbation manifold, which grows in dimension with each joint made available to the system.

We compare two platforms. A baseline platform has access to its kinematic chain, so induced disturbances on the body cannot be accounted for:
\begin{equation}
\label{eq:kin}
\dot{x}_{ee} = J(\theta)\dot{\theta}.
\end{equation}
It takes a pseudoinverse of the Jacobian to compute $\dot{\theta}$ given a desired velocity for the end effector $\dot{x}_{ee}$.
The second platform has access to both its kinematic chain and the model of the viscous fluidic model that it inhabits:
\begin{equation}
\label{eq:dynpfaffian}
\begin{bmatrix} \dot{x}_{ee} \\ \dot{x}_{b} \end{bmatrix} = \begin{bmatrix} J(\theta) \\ P(\theta) \end{bmatrix} \dot{\theta}.
\end{equation}
where $\dot{x}_{b}$ can be set to zero.

This toy system is a simple adaptation of the Purcell swimming model \cite{becker2003self,hatton2013geometric}, where we magnify the ratio of across-to-along body forces by a factor of 5, mimicking a more fin-like relationship with the water to excite larger perturbations.
In the trial, $\dot{x}_{ee}$ is set by the residual between the location of the hand (with respect to the base) and the desired hand location (with respect to the base). 
The desired hand location is a function of time that traces each shape (from right to left)). 
The closest hand is driven to track the desired hand location, the other being free to use its neighboring links as a tail-like appendage if necessary to stabilize the base.
Fig.~1 shows that the thirteen-link swimmer was capable of exactly tracing the desired shapes with a closed form solution when leveraging the  pfaffian form of the dynamics. 
Meanwhile, the platform embedded with only kinematic-chain knowledge both drifted and was unable to draw shapes resembling those desired.
These results were not intended to capture phenomena such as collision-avoidance including self-collisions, but these topics are addressed in the next section.

\section{RRT on Zero Perturbation Manifolds}
\label{sec:method}

In Sections~\ref{sec:bg} and \ref{sec:zpm} we overviewed a key class of floating-based dynamics and its relevance for distinction of a zero-perturbation manifold that occupies $TB$.
Here, we will show how such a manifold can be used effectively within sampling-based motion planners for rapid exploration based planning on high degree of freedom floating-base robots in obstacle rich settings.
Equation~(\ref{eq:hard_to_connect}) shows us that connecting two points in the shape space can be challenging.
This suggests that objects like probabilistic roadmaps (PRMs) will be structurally challenging to generate, since the inter-connectedness of nodes requires direct confrontation of this issue.
Turning our attention to single-query planning approaches, RRT offers the ability to grow quickly from a known location.
As we have established, knowing the final location does not make it easy to find a plan that arrives exactly at the final location.
One might think of a car placed at a close distance but incorrect orientation to a desired pose. 
While Lie brackets or primitive-based motion planning may offer a solution for the car, it is not obvious that such approaches are computationally tractable in the floating-base manipulation setting.

We observe three options, which include growing the RRT from an initial location to a desired location, growing a bi-RRT from both locations towards each other, and growing from the desired location toward the initial location.
We accept the third option as suboptimal but highly practical.
An RRT will be designed to rapidly grow from a location of interest toward the robots initial configuration, all the while restricting itself to the zero perturbation manifold.
The optimal path selected from this tree is the core contribution of this work. 
The robot will start by connecting itself from initial configuration to the ZPM. Once it reaches the trajectory along the ZPM, we assert that it will stop while the thrusters mitigate any induced perturbation.
Now, after regulation, the floating-base system is afforded a zero perturbation path, through clutter, to the object of interest.

\begin{algorithm}
\caption{$ZPMRRT(\theta_s,\theta_g)$}
\begin{algorithmic}[1]
\label{alg:main}
\State T.init($\theta_g$) 
\While{$TimeRemaining()$}
    \State $\theta_{sample} \gets sample(i,\theta_s)$ 
    \State $\theta_{nearest} \gets nearest(T,\theta_{sample})$
    \State $\theta_{extend} \gets ExtendAlongZPM(T,\theta_{nearest},\theta_{sample})$
    \If{$||\theta_{extend}-\theta_s|| < \alpha$} 
    \State \textbf{return} $Smooth(ExtractPath(T,\theta_s))$
    \EndIf 
\State \textbf{return} $null$
\EndWhile
\end{algorithmic}
\end{algorithm}
    
\begin{algorithm}
\caption{$ExtendAlongZPM(T,\theta_{near},\theta_{sample})$}
\begin{algorithmic}[1]
\label{alg:ExtendAlongZPM}
\State $\theta_{extend} \gets \theta_{near}; \texttt{ } spacing \gets 0$
\While{$Timeout()$ \textbf{ or } $Collision(\theta_{extend})$}
    \State $\delta\theta \gets \theta_{extend}-\theta_{sample}$
    \State $u,s,v \gets svd(P(\theta_{extend}))$
    \State $P_{null} \gets v[dim(G):]$
    \If{$P_{null} \perp \delta\theta$}
        \State \textbf{return}  $\theta_{extend}$
    \Else
        \State $\theta_{extend} \gets \theta_{extend} + unit(\delta\theta P_{null}^T P_{null})dt$
        \State $spacing \gets spacing + dt$
        \If{$spacing > L$}
            \State $T.addVertex(\theta_{near},\theta_{extend})$
            \State $\theta_{near} \gets \theta_{extend}; \texttt{ } spacing \gets 0$
        \EndIf
    \EndIf
\EndWhile
\State \textbf{return} $\theta_{extend}$
\end{algorithmic}
\end{algorithm}

\label{sec:results}
\begin{figure*}
\label{fig:zpmrrt_rrt_cartoon}
    \includegraphics[width=\textwidth]{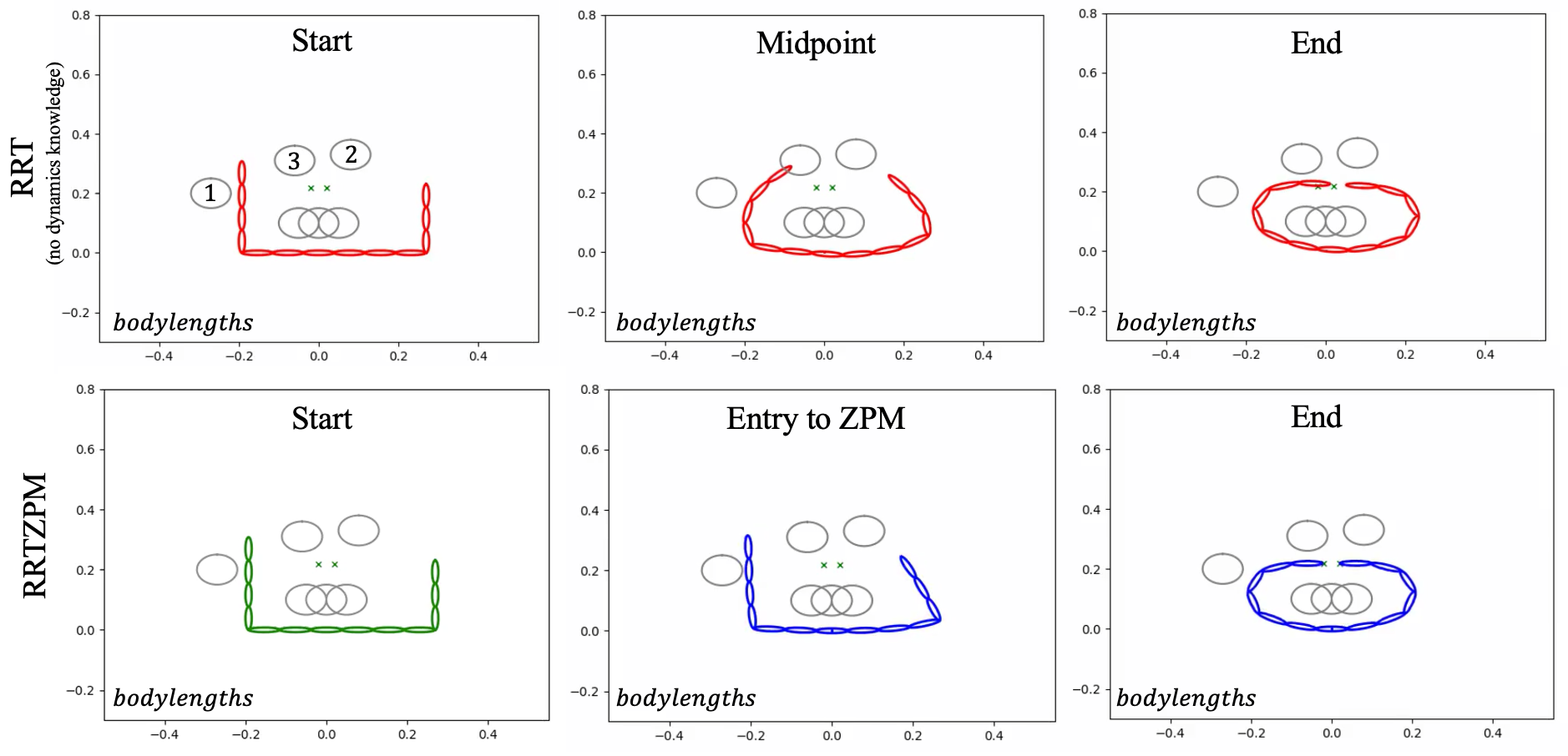}
    \caption{Using a classical RRT (top), the thirteen-link Purcell swimmer is not able to accurately obtain a desired pose during emplacement (taken from 1 of 30 trials detailed in Fig.~3). We show snapshots of the swimmer at the beginning (first column), midpoint (second column), and end (final column), approaching hand locations (pairs of green x's) while avoiding obstacles (grey circles). Using ZPMRRT (bottom), the swimmer is able to reach the desired endpoint. We show snapshots of the behavior, with columns carrying the same timestamps with the exception of the middle, which marks the transition point to the ZPM. Accompanying video in Supplementary Materials. }
\end{figure*}

\begin{figure}[H]
\label{fig:zpmrrt_rrt_trials}
    \includegraphics[width=.49\textwidth]{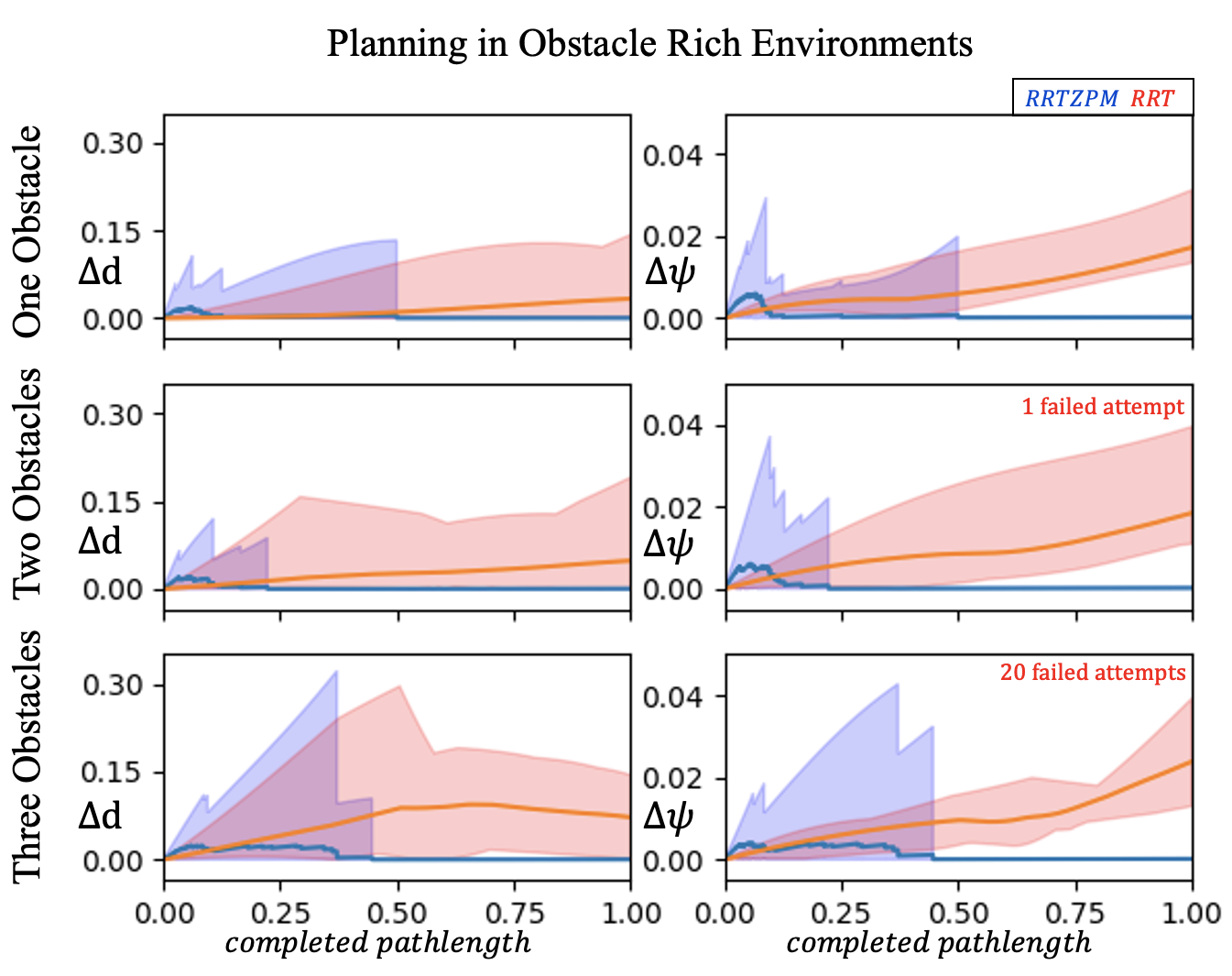}
    \caption{This Fig.~show the displacement of the Purcell swimmer in distance (left $\Delta d$ in bodylengths) and orientation (right $\Delta \psi$ in radians) over 30 trials for emplacement with one obstacle (top row, obstacle numbers identified in Fig.~2), two obstacles (middle row), and three obstacles (bottom row). The mean (solid) and bounds (transparent band) of performance are plotted for the RRTZPM (blue, Algorithm~1) motivated in this work as well as the classical RRT (red) which does not incorporate any dynamics knowledge.}
\end{figure}

Here we will outline our approach for floating-base manipulation along zero perturbation manifolds.
Algorithm~1 is provided a start and goal internal configuration for the robots ($\theta_s,\theta_g$). 
As described above, we will start at the goal location, where there is a higher need for precision (such as in emplacement and extraction tasking). 
We initialize a tree $T$ with this node and query a sample function at each iteration. This will provide a random internal configuration or the start configuration (with probability $0.5$ in this work).
We then attempt to grow the tree toward this sampled configuration, while adhering to the zero perturbation manifold. 
The computation details of this process are available in Algorithm~2, which involves computation of the null space via singular value decomposition (SVD) and projection of the error between goal and current point along the computed subspace of the internal configuration that carries zero eigenvalues. 
If we avoid singular configurations (as is assumed in this algorithm), then the null space is consistently of $rank(null(P))=dim(B)-dim(G)$ and can be indexed directly as the corresponding bottom rows of V.
We can project the growth direction onto this manifold (see line 9 of Algorithm~2 where $unit$ produces a unit vector), and stop extending in the event of a timeout, collision, or perpendicular orientation of the null space and remaining error.
During extension, we grow the tree by adding a node after such a length has been extended from a prior node in the tree.
Once the search has timed out or achieved sufficient proximity to the goal configuration, the algorithm returns a smoothed path.
In this work, smoothing involves attempting to find a shortcut between the start and end point of the path by direct projection onto the ZPM, then recursively repeating the procedure on two randomly split segments of the path (until those segments reduce to two points). 
This smoothing process is repeated three times in this work.

\section{Results}

Here we will ask the thirteen-link Purcell swimmer to plan in constrained environments while executing an emplacement task. 
This task will require that both hands reach a particular location in the workspace. 
Through expectation that the base does not move, we stipulate a shape that will achieve the desired end effector locations and provide this desired shape as a goal to the planner.
As mentioned before, the planner will grow from the goal towards the starting location. 
We identify that there are a variety of viable starting configurations from which to approach the emplacement task, and encode this within the planner.
When trying to connect to starting shapes $\theta_s$, we will provide a randomly selected, generalized $U$-shape for the swimmer. 
\footnote{A $U$ includes a negative quarter rotation of a joint left of the base and a positive quarter rotation of a joint right of the base.}
We can provide this freedom of choice since we will assume that the system can start from a further displaced position, change configuration to any other $U$-shape, and then trivially re-engage at the position identified in the planner\footnote{Any such shape can translate in the coordinate perpendicular to the base link to exit or enter the starting configuration, making it a viable entry point to start the emplacement plan from.}.
Thus, we claim to be solving a broader planning problem (starting from a disengaged, distal position with any $U$-shaped initial joint configuration) even though we only explicitly solve the non-trivial component of the planning problem (once engaged in an obstacle rich environment).

We start with one obstacle and provide the RRTZPM 30 separate trials to find connections to $U$-shapes along the zero perturbation manifolds. 
We then ask the RRT to connect the same two configurations.
In each trial we allow the planner to replan 4 times after 10s intervals if a collision free path is not found. 
Paths are termed collision free if they would not cause a collision with a static base.
In every trial both planners succeed during the first 10s interval. 
Fig.~3 shows the displacement as a function of a fraction of path length, spanning the start and finish of the attempt. 
It is clear that the proposed method accrues a brief initial displacement to access the zero peturbation manifold (distinguishing the effect of the properties following from (\ref{eq:hard_to_connect}).
However, once on the ZPM, the platform is capable of achieving the desired configuration with zero displacement.
Meanwhile, the classical RRT accrues error, critically during the final steps of engagement of the desired internal configuration. This classical approach persistently accrues error as it approaches the final location, exactly when the proposed method accrues zero error by structure.

We see this relationship persist as a second and third obstacle are added. Notably, during trials with 3 obstacles, the RRT unsuccessfully attempts to find a collision free path in 20 of 30 trials.
Across all numbers of obstacles, the proposed method (for each obstacle) experiences an initial perturbation of the base followed by an approximately zero perturbation approach to the final configuration.
We observe this difference in performance in Fig.~2, where we can see the drift over the duration of the trial for the classical RRT on the top row, as well as the lack of drift for the example on the bottom row.

\begin{figure*}
\label{fig:zpmrrt_lqr_cartoon}
    \includegraphics[width=\textwidth]{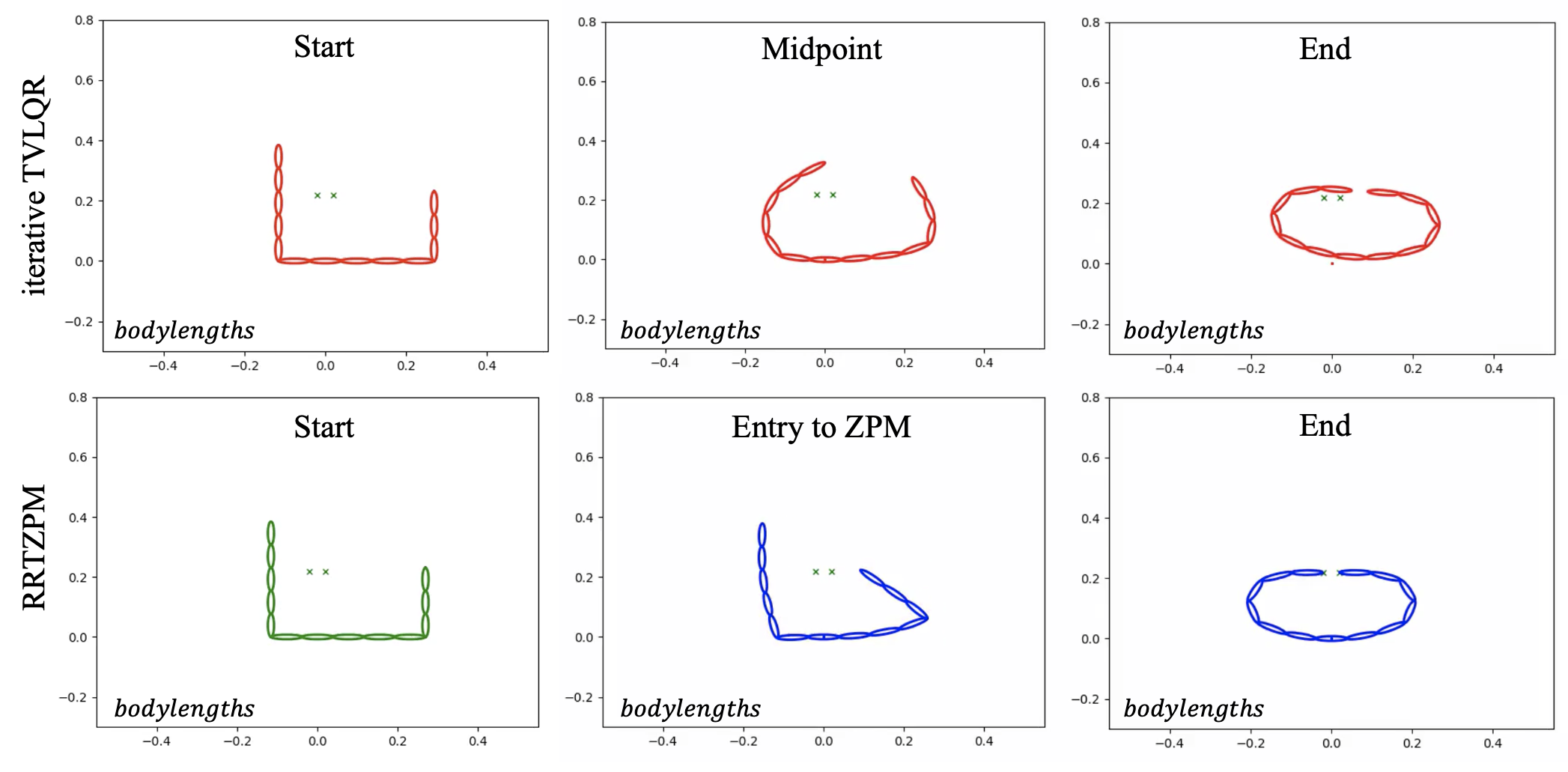}
    \caption{Using iterative TVLQR (top), the thirteen-link Purcell swimmer is not able to accurately obtain a desired pose during emplacement (taken from 1 of 30 trials in Fig.~5). We show snapshots of the swimmer at the beginning (first column), midpoint (second column), and end (final column), approaching hand locations (pairs of green x's) while avoiding obstacles (grey circles). Using RRTZPM (bottom), the swimmer is able to reach the desired endpoint. We show snapshots of the behavior, with columns carrying the same timestamps with the exception of the middle, which marks the transition point to the ZPM. Accompanying video in Supplementary Materials. }
\end{figure*}

\begin{figure}
\label{fig:zpmrrt_lqr_trials}
    \includegraphics[width=.49\textwidth]{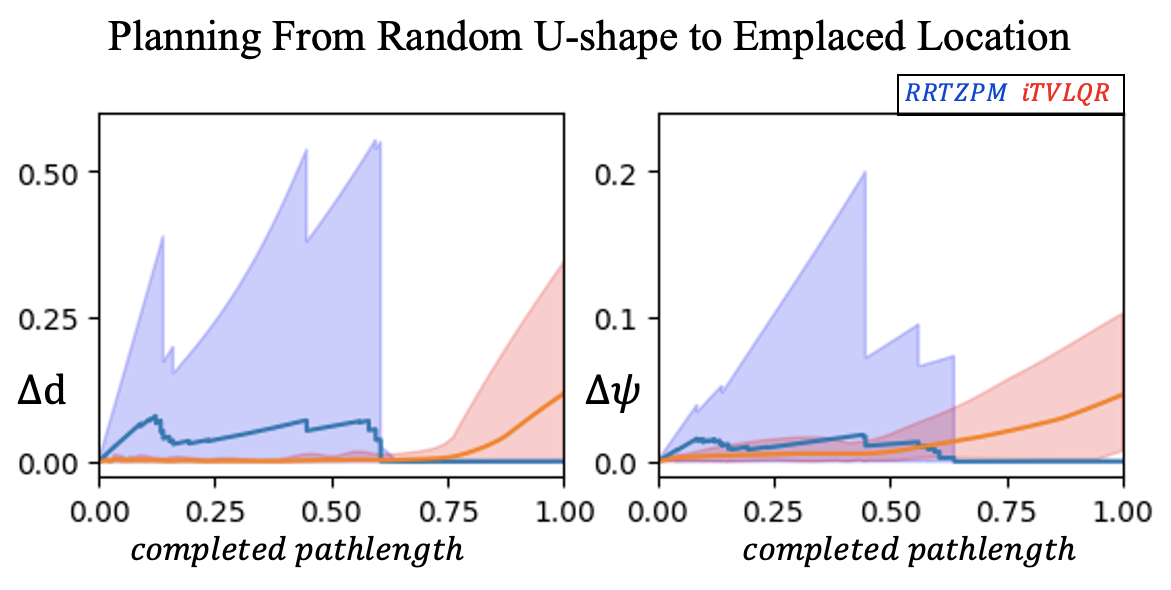}
    \caption{This Fig. shows the displacement of the purcell swimmer in distance (left $\Delta d$ in bodylengths) and orientation (right $\Delta \psi$ in radians) over 30 trials with no obstacles. The mean (solid) and bounds (transparent band) of performance are plotted for RRTZPM (blue, Algorithm~1) motivated in this work as well as the iterative TVLQR which is subject to the locality of exploitation-driven solutions.}
\end{figure}

In obstacle-free environments we compare this approach to an iterative time varying linear quadratic regulator (TVLQR). 
This provides the opportunity to observe the exploration-driven solutions available via RRTZPM with respect to this exploitation-driven approximation of the Hamilton-Jacobi-Bellman equation. 
The iterative TVLQR is seeded with a direct route from initial to final shape for the linear system:
\begin{equation}
\begin{bmatrix} \dot{x}_{b} \\ \dot{\theta} \end{bmatrix} = \begin{bmatrix}  P(\theta) \\ I \end{bmatrix} \dot{\theta}.
\end{equation}
where we take the shape velocity $\dot{\theta}$ as the control input. 
We discretize each trajectory at 50 samples. For $n=dim(B)$ and $m=dim(G)$ take 
\begin{align}
Q=\begin{bmatrix} I^{mxm} && 0 \\ 0 && 0 \end{bmatrix} && Q_f=\begin{bmatrix} I^{mxm} && 0 \\ 0 && I^{nxn} \end{bmatrix} && R=Ie^{-2}
\end{align}
where $Q_f$ is the final element of $Q$.
By iteratively supplying the output of a TVLQR solution to itself, we provide the opportunity to further refine the solution with the context of the dynamics centered at the most recent solution. 
At the end of 20 iterations we have a trajectory that has been locally refined to find the desired joint location while avoiding perturbations.
We ask the swimmer to achieve the previously identified emplacement shape configuration from randomly selected $U$-shapes (this time the swimmer cannot select a $U$-shape.) 
Fig. 4 shows that over 30 trials we observe the RRTZPM is capable of accessing the ZPM, although it accessed later in the maneuver than in Fig~3 when it had the choice of starting $U-shape$ locations. 
The iterative TVLQR critically accrues error as it arrives at the emplacement location, repeating the precision mitigating behavior seen in the RRT in Fig. 3. 
We can see impact of this error at the emplacement location in Fig.~5, where the error accrued through iterative TVLQR renders the platform unable to complete the emplacment task.

\section{Discussion}
\label{sec:discussion}

The active use of the other arm to decouple the generated drag force in Fig.~1 reminded us of the use of tails in many animals.
This may provide inspiration for design of systems in similar dynamic environments.
Overall, the end effector tracking results of Fig.~1 are not entirely surprising (prior work exists in null space projection for swimming robots \cite{brantner2021controlling}). 
The example reinforces what is known in manipulation, that redundancy offers a greater variety of options to achieve the desired result.
Observing this property in the dynamic case while tracking shapes informs us that this property, since it exists for a local tracking problem, may be possible to embed in a more global planning problem with more constraints.

The emplacement performance of the RRTZPM method in Fig.~3 and Fig.~2 shows us that it was possible to embed this property in a high degree of freedom platform with three obstacles. It is clear from comparison to the RRT planner (that neglects dynamics) that the ZPM is providing significant value. The practical relevance of this result is made clear in Fig.~2, where the RRT platform can be seen colliding and drifting from the desired location. Such precision is unacceptable for even coarse emplacement tasking.

This relationship persists when we relate the motivated method to iterative TVLQR in an obstacle-free environment. The exploitation driven algorithm can locally refine the initial trajectory, but is unable to find a more global solution that permits acceptable precision for the emplacement task. This supports the perceived value of our method which can respect dynamics while finding inventive solutions rapidly.

The measured perturbation rejection on these simulated systems can be impacted largely by the time-step of the integration. In principle, simulating at a small enough time step would permit a numerically zero perturbation when integrating along the ZPM. For real systems, feasibility of precisely tracking shape evolution along the ZPM may significantly impact the quality of the overall plan.

\section{Conclusion}
\label{sec:conclusion}

The results suggest that the proposed methods provide a novel class of algorithms that can address the weaknesses of classical sampling-based motion planners and optimization-based approaches.
We showed that we were able to maintain the "sampling-scalability at high dimension" feature of RRTs while providing them the dynamical context to avoid precision-related failures. 
We showed that the proposed method maintained the exploration-geared inventiveness of RRTs, permitting viable solutions where exploitation-driven optimization schemes will fail.

Our examples are restricted to a planar swimming simulated system. Further questions will remain pertinent in extensions of these results to other systems, simulated or real. 
How viable are pfaffian constraint approximations for systems that do not exist exactly in high or low Reynolds domains? 
How many degrees of freedom are typically needed to achieve similar results in three-dimensional scenarios? 
How easy is it to execute trajectories along this manifold with conventional hardware?
In future work we plan to explore open questions directly on hardware in underwater and orbital domains.

\printbibliography

\end{document}